\documentclass[runningheads]{llncs}
\usepackage{graphicx}

\usepackage{tikz}
\usepackage{comment}
\usepackage{amsmath,amssymb} 
\usepackage{color}
\usepackage{booktabs}
\usepackage{multirow}
\usepackage{wrapfig}
\usepackage{enumitem}
\usepackage{caption}
\usepackage{diagbox}
\usepackage{MnSymbol}
\usepackage{blindtext}
\usepackage[symbol]{footmisc}

\usepackage[accsupp]{axessibility}  

\begin{document}
\pagestyle{headings}
\mainmatter
\def\ECCVSubNumber{1493}  

\title{Quantized GAN for Complex \\ Music Generation from Dance Videos} 

\titlerunning{D2M-GAN for Dance Music Generation}
%
\author{Ye Zhu\inst{1}\thanks{This work was mainly done while the author was an intern at Snap Inc.}\and Kyle Olszewski\inst{2} \and Yu Wu\inst{3} \and Panos Achlioptas\inst{2} \and Menglei Chai\inst{2} \and \\
Yan Yan\inst{1} \and Sergey Tulyakov\inst{2}}
\authorrunning{Y. Zhu et al.}
%
\institute{Illinois Institute of Technology, USA \\
\and
Snap Inc., USA \\
\and
Princeton University, USA 
}

\maketitle
\begin{abstract}
We present Dance2Music-GAN (D2M-GAN), a novel adversarial multi-modal framework that generates complex musical samples conditioned on dance videos. Our proposed framework takes dance video frames and human body motions as input, and learns to generate music samples that plausibly accompany the corresponding input.
Unlike most existing conditional music generation works that generate specific types of mono-instrumental sounds using symbolic audio representations (\textit{e.g.}, MIDI), and that usually rely on pre-defined musical synthesizers, in this work we generate dance music in complex styles (\textit{e.g.}, pop, breaking, etc.) by employing a Vector Quantized (VQ) audio representation, and leverage both its generality and high abstraction capacity of its symbolic and continuous counterparts.
By performing an extensive set of experiments on multiple datasets, and following a comprehensive evaluation protocol, we assess the generative qualities of our proposal against alternatives. The attained quantitative results, which measure the music consistency, beats correspondence, and music diversity, demonstrate the effectiveness of our proposed method. Last but not least, we curate a challenging dance-music dataset of in-the-wild TikTok videos, which we use to further demonstrate the efficacy of our approach in \textit{real-world} applications -- and which we hope to serve as a starting point for relevant future research. Dataset and code at \url{https://github.com/L-YeZhu/D2M-GAN}.
\end{abstract}

\section{Introduction}

\textit{``When the music and dance create with accord, their magic captivates both the heart and the mind."}~\footnote[2]{Jean-Georges Noverre.}
As a natural form of expressive art, dance and music have enriched our daily lives with a harmonious interplay of melodies, rhythms, and movements, across the millennia.
The growing popularity of social media platforms for sharing dance videos such as TikTok has also demonstrated their significance as a source of entertainment in our modern society. 
At the same time, new research works are flourishing following the trend and exploring multi-modal generative tasks between dance motions and music~\cite{lee2019dancing,aistplus,li2021dancenet3d,aggarwal2021dance2music}.

Although seemingly intuitive, music generation from dance videos has been a challenging task due to two main reasons. First, typical audio music signals are high-dimensional and require sophisticated temporal correlations for overall coherence~\cite{review1_ji2020comprehensive,review2_briot2020deep}. For example, CD-quality audio has a typical sampling rate of 44.1 kHz, resulting in over 2.5 million data points (``dimensions") for a one-minute musical piece~\cite{jukebox}.
In contrast, most dance generation works output the relatively low-dimensional motion data in the form of 2D or 3D skeleton keypoint (\textit{e.g.}, displacement for dozens of joints) conditioned on the music~\cite{lee2019dancing,aistplus,am1-shlizerman2018audio,am2-ren2020self}, which are then rendered into dance sequences and videos. 
To tackle the challenge of the high dimensionality of audio data, the research studies on music generation from visual input~\cite{foley,su2020audeo,musictransformer} often rely on the low-dimensional intermediate symbolic audio representations (\textit{e.g.}, 1D piano-roll or 2D MIDI). The symbolic representations benefit existing learning frameworks with a more explicit audio-visual correlation mapping and more stable training, as well as widely-established standard music synthesizers for decoding the intermediate representations. 
However, such symbolic-based works suffer from the limitations on the flexibility of the generated music, which brings us to the second challenge of dance video conditioned music generation. 
Specifically, a separately trained model is usually required for \textit{each} instrument and the generated music is composed with acoustic sounds from a \textit{single predefined} instrument~\cite{foley,musegan,midi-event} (\textit{e.g.}, imagine a person dancing hip-hop with piano-based music). These facts make existing conditional music generation works difficult to generalize in complex musical styles and real-world scenarios.

\begin{figure}[t]\small
    \centering
    \includegraphics[width=0.63\textwidth]{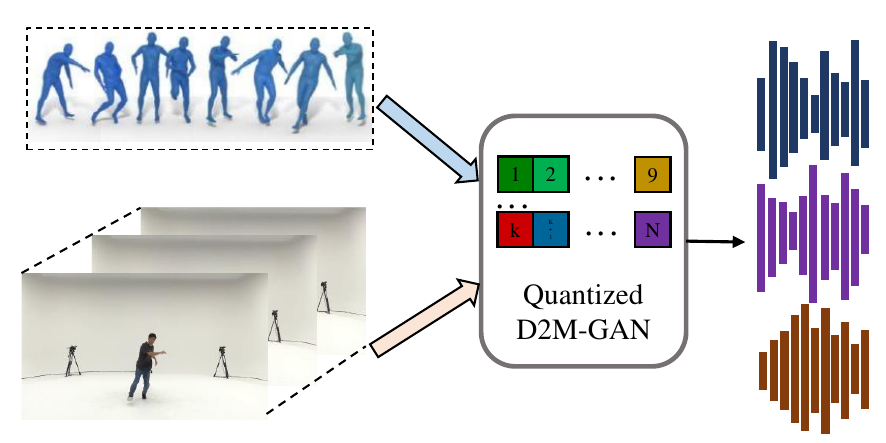}
    \caption{\textbf{Task illustration.} We introduce a Vector-Quantized-based framework for music generation from dance videos, which takes human body motions and visual frames as input, and generates suitable music accordingly. Our proposed model is able to generate complex and rich dance music - in contrast to most existing conditional music generation works that typically output mono-instrumental sounds.}
    \label{fig1:task}
\end{figure}

To fill this gap, we propose a novel adversarial multi-modal framework that learns to generate complex musical samples from dance videos via the Vector Quantized audio representations. 
Inspired by the recent successes of VQ-VAE~\cite{vqvae,vqvae2,jukebox} and VQ-GAN~\cite{vqgan2021}, we adopt quantized vectors as our intermediate audio representation, and leverage both their increased abstraction ability compared to continuous raw audio signals, as well as their flexibility of better representing complex real-world music compared to classic symbolic representations.
Specifically, our framework takes the visual frames and dance motions as input (Fig.~\ref{fig1:task}), which are encoded and fused to generate the corresponding audio VQ representations. After a lookup process of the generated VQ representations in a learned codebook, the retrieved codebook entries are decoded back to the raw audio domains using a fine-tuned decoder from JukeBox~\cite{jukebox}.
Additionally, we deploy a convolution-based backbone and follow a hierarchical structure with two separate abstraction levels (\textit{i.e.}, different hop-lengths) for the audio signals to test the scalability of our framework. The higher-level model has a larger hop-length and fewer parameters, resulting in faster inference. In contrast, the lower-level model has a lower abstraction level with smaller hop-length, which enables the generation of music with higher fidelity and better quality. 

Last but not least, we also contribute a real-world paired dance-music dataset collected from TikTok video compilations. Our dataset contains in total 445 dance videos with 85 songs and an average per-video duration of approximately 12.5 seconds. Unlike existing datasets (e.g., AIST~\cite{aist-dance-db,aistplus}), ours is more challenging and better reflects the conditions of real-world scenarios, setting thus a new point for relevant future research.

Tapping on such datasets, we conduct extensive experiments to demonstrate the effectiveness and robustness of the proposed framework. Specifically, we design and follow a rich evaluation protocol to consider its generative quality with respect to the correspondence of dance input in in terms of beats, genres and coherence, the general quality of the generated music is also assessed. The attained results (both quantitative and qualitative) show that our model can generate plausible dance music in terms of various musical features, outperforming several competing conditioned music generation methods.

In summary, our main contributions are:
\begin{itemize}[leftmargin=*, topsep=1pt, partopsep=0pt, itemsep=1pt]
    \item We propose \emph{D2M-GAN}, a novel adversarial multi-modal framework that generates complex and free-form music from dance videos via \textit{Vector Quantized (VQ) representations}.
    
    \item Specifically, the proposed model, using a VQ generator and a multi-scale discriminator, is able to effectively capture the temporal correlations and rhythm for the musical sequence to generate complex music. 

    \item To assess our model we introduce a comprehensive \textit{evaluation protocol} for conditionally generated music and demonstrate how the proposed \emph{D2M-GAN} is able to generate more complex and plausible accompanying music compared to existing approaches.
    
    \item Last but not least, we create a novel real-world dataset with dance videos captured \textit{in the wild} -- and use it to establish a new more challenging setup for conditioned music generation, which further demonstrates the superiority of our framework.

\end{itemize}

\section{Related Work}
\label{sec:related_work}

\noindent \textbf{Audio, Vision and Motion.}
Combining data from audio, vision, and motion has been a popular research topic in recent years within the field of multi-modal learning~\cite{zhu2022skeleton,wu2022snoc,zhu2021learning,am2-ren2020self,am3-ferreira2021learning}.
Research focusing on general audio visual learning typically assumes that the two modalities are intrinsically correlated based on the synchronization nature of the audio and visual signals~\cite{av1-2018cooperative,av2-2018audio,av3-2016ambient,zhu2021learning,av4-2017look,av5-2016soundnet}. Such jointly learned audio-visual representations thus can be applied in multiple downstream tasks like sound source separation~\cite{sss1-2020music,sss2-2018learning,sss3-20192,sss4-gao2019co,sss5-zhao2019sound}, audio-visual captioning~\cite{avcaption1-2019watch,accaption2-2018watch}, audio-visual action recognition~\cite{gao2020listen,kazakos2019epic}, and audio-visual event localization and parsing~\cite{tian2018audio,wu2019dual,zhu2021learning,wu2021explore}.

On the other hand, another branch of studies related to our work has been investigating the correlations between motions and sounds~\cite{lee2019dancing,zhuang2020music2dance,aistplus,foley,li2021dancenet3d}. A large portion of the research works here, aim to generate human motions based on the audio signals, either in the form of 2D pose skeletons~\cite{am1-shlizerman2018audio,am2-ren2020self,lee2019dancing} or direct 3D motions~\cite{aistplus,am4-tang2018dance,am5-kao2020temporally}. For the inverse direction that seeks to generate audio from motions, Zhao \textit{et al.}~\cite{sss5-zhao2019sound} introduces an end-to-end model to generate sounds from motion trajectories.
Gan \textit{et al.}~\cite{foley} propose a graph-based transformer framework to generate music from performance videos using raw movements as input. 
Di \textit{et al.}~\cite{di2021video} propose to generate video background music conditioned on the motion and special timing/rhythmic features of the input videos. 
In contrast to these previous works, our work combines three modalities, which takes the vision and motion data as input and generates music accordingly. 

\noindent \textbf{Music Generation.}
Raw music generation is a challenging task due to the high dimensionality of the audio data and sophisticated temporal correlations. Therefore, the existing music generation approaches usually adopt an intermediate audio representation for learning the generative models to reduce the computational demand and simplify the learning task~\cite{melgan,musegan,musictransformer,jukebox,wavenet,zhu2022discrete}. Classic audio representations mainly include the symbolic and continuous categories.
Musegan~\cite{musegan} introduces a multi-track GAN-based model for instrumental music generation via the 1D piano-roll symbolic representations.
Music Transformer~\cite{musictransformer} aims to improve the long-term coherence of generated musical pieces using the 2D event-based MIDI-like audio representations~\cite{midi-event}.
Melgan~\cite{melgan} is a generative model for music in form of the audio mel-spectrogram features.  
Recently, JukeBox~\cite{jukebox} introduces a generic music generation model based on the novel Vector Quantized (VQ) representations.
Our proposed framework adopts the VQ representations for music generation.

\noindent \textbf{Vector Quantized Generative Models.}
VQ-VAEs~\cite{vqvae,vqvae2} are firstly proposed as a variant of the Variational Auto-Encoder (VAE)~\cite{vae-kingma2013auto} with discrete codes and learned priors. Following works have demonstrated the potential of VQ-based framework in multiple generative tasks such as image and audio synthesis~\cite{vqgan2021,jukebox,SpecVQGAN_Iashin_2021}. Specifically, the VQ-VAE~\cite{vqvae} is initially tested for generating images, videos, and speech. An improved version of VQ-VAE~\cite{vqvae2} is proposed with a multi-scale hierarchical organization. Esser \textit{et al.}~\cite{vqgan2021} apply the VQ representations in the GAN-based framework for generating high-resolution images. Dhariwal \textit{et al.}~\cite{jukebox} introduce the JukeBox as a large-scale generative model for music synthesis based on VQ-VAE. 
Compared to the symbolic and continuous representations, VQ representations leverage the benefits of flexibility (\textit{i.e.}, the ability to represent complex music genres with a unified codebook in contrast to symbolic representations) and high compression levels (\textit{i.e.}, the learned codebooks largely reduce the data dimensionality compared to raw waveform or spectrogram). Our proposed framework combines both the GAN~\cite{gan} and VAE~\cite{vae-kingma2013auto}, which uses the GAN-based learning to generate VQ representations from the dance videos, and adopts the VAE-based decoder for synthesizing music.

\section{Method}

\begin{figure}[t]\small
    \centering
    \includegraphics[width=1.0\textwidth]{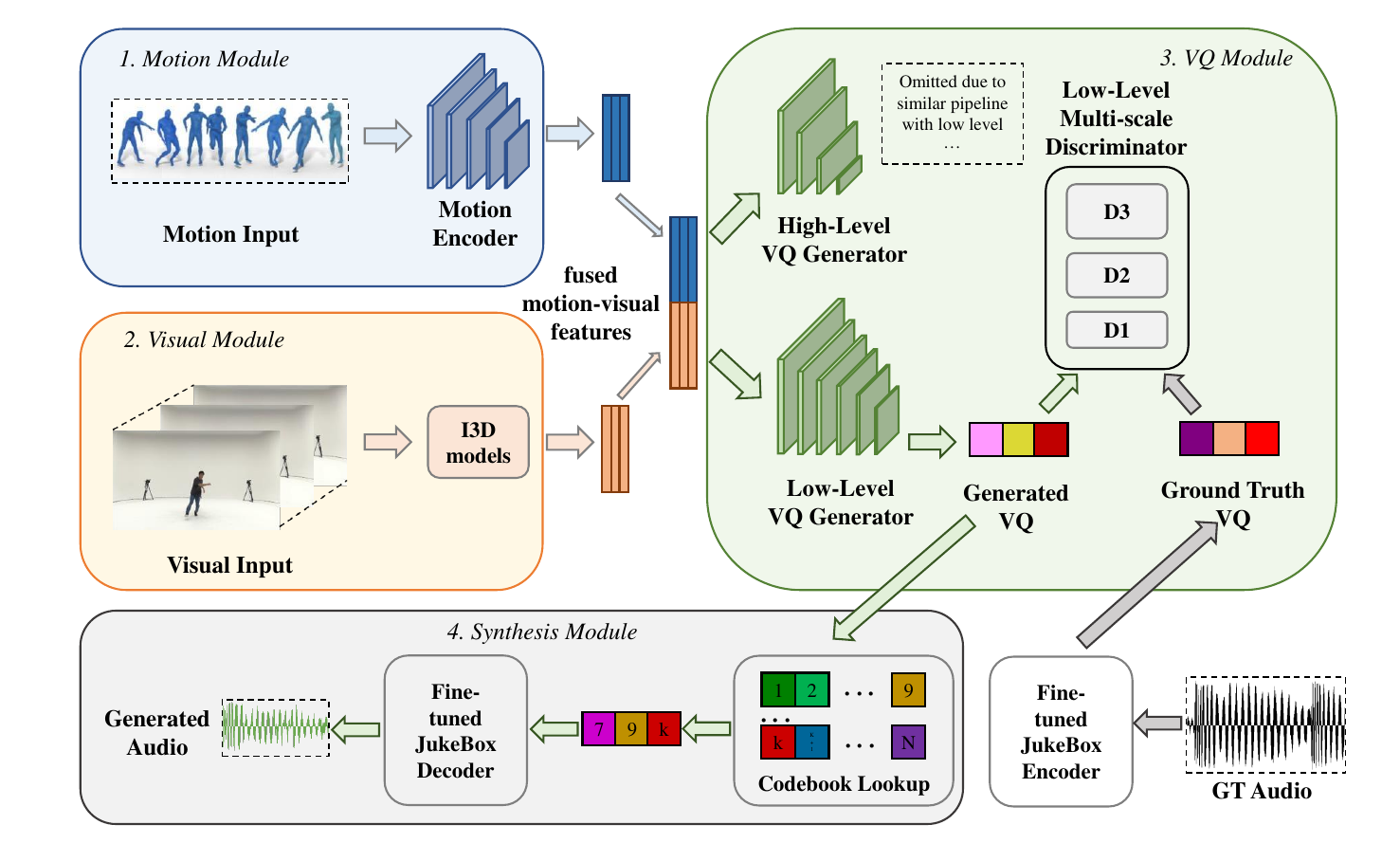}
    \caption{\textbf{Overview of the proposed architecture of the \emph{D2M-GAN}}. Our model takes the motion and visual data from the dance videos as input and process them with the motion and visual modules, respectively. It then forwards the concatenated representation containing information from both modalities to ground the generation of audio VQ-based representations with the VQ module. The resulting features are calibrated by a multi-scale GAN-based discriminator and are used to perform a \textit{lookup} in the pre-learned codebook. Last, the retrieved codebook entries are decoded to raw musical samples via by a pre-trained and fine-tuned decoder, responsible for synthesizing music.}
    \label{fig2:overall}
\end{figure}

An overview of the architecture of the proposed D2M-GAN is shown in Fig.~\ref{fig2:overall}. Our approach entails a hierarchical structure with two levels of models that are independently trained with a similar pipeline for flexible scalability. For each level, the model consists of four modules: the motion module, the visual module, the VQ module consisting of a VQ generator and the multi-scale discriminators, and the music synthesizer. Our hierarchical structure amplifies the flexibility to choose between the trade-off of the music quality and computational costs according to practical application scenarios. A detailed description of these modules is given below while further architectural details and model-selection-tuning are included in the supplementary.

\subsection{Data Representations}

During the inference, the input to our proposed \emph{D2M-GAN} come from two major modalities: the visual frames of the dance videos and human body motions of dance performers. The ground-truth music audio is also used as the supervision for the discriminators during the training stage. 
For the human body motions, several different forms of data representations such as 3D Skinned Multi-Person Linear model (SMPL)~\cite{SMPL:2015} or 2D body keypoints~\cite{openpose-cao2017realtime,openpose} can be utilized by our framework. We use SMPL and 2D body keypoints for different datasets in our experiments.
To encode the visual frames, we extract I3D features~\cite{i3d} using a model pre-trained on Kinectics~\cite{kay2017kinetics}.
For the musical data, we adopt the VQ as the intermediate audio representation. To leverage the strong representation ability of codebooks trained on the large-scale musical dataset, we use the pre-learned codebooks from JukeBox~\cite{jukebox}, which are trained on a dataset of 1.2 million songs.

\subsection{Generator}

The generator $G = \{G_m, G_v, G_{\mathrm{vq}}\}$ includes the motion module $G_m$, the visual module $G_v$, and the principal VQ generator $G_{vq}$ in the VQ module, which takes the fused motion-visual data as input and outputs the desired VQ audio representations. 
\begin{equation}
    \centering
    f_{\mathrm{vq}} = G_{\mathrm{vq}}(G_m(x_m),G_v(x_v)) = G(x_m, x_v),
\end{equation}
where $x_m$ and $x_v$ represent the motion and visual input data, respectively. $f_{\mathrm{vq}}$ is the output VQ representations.
All these modules are implemented as convolution-based feed-forward networks. For the principal VQ generator, we use leaky rectified activation functions~\cite{leaky-xu2015empirical} for its hidden layers and a tanh activation for its last layer before output to promote the stability of GAN training~\cite{radford2015unsupervised}.

It is also worth noting that we find that using batch normalization and the aforementioned activation function designs~\cite{lucic2017gans,radford2015unsupervised,salimans2016weight} is crucial for a stable GAN training in our framework. However, the application of the tanh activation will also restrict the output VQ representations within the data range between $-1$ and $+1$. We choose to scale activation after the last tanh activation by multiplying by a factor $\sigma$. The hyper-parameter $\sigma$ enlarges the data range of VQ output and makes it possible to perform the lookup of pre-learned large-scale codebooks $\mathrm{LookUp}(f'_{\mathrm{vq}})$ with $f'_{\mathrm{vq}} = \sigma f_{\mathrm{vq}}$.
Another significant observation regarding the generator's design is using a wide receptive field. Music has long temporal dependencies and correlations compared to images, therefore, the principal VQ generator with a larger receptive field is beneficial for generating music samples with better quality, which is consistent with the findings from previous works~\cite{melgan,donahue2018adversarial}. To this end, we design our generator with relatively large kernel sizes in the convolutional layers, 
and we also add residual blocks with dilations after the convolutional layers. All previously described sub-modules within our generator $G$ are jointly optimized.

\subsection{Multi-Scale Discriminator}
Similar to the generator, the discriminator in the D2M-GAN is also expected to capture the long-term dependencies of musical signals encoded in the generated sequence of VQ features.
However, different from the generator design that focuses on increasing the receptive fields of the neural networks, we address this problem in the discriminator design by using a multi-scale architecture. 
\begin{wrapfigure}[20]{r}{0.35\textwidth} \small
    \centering
    \includegraphics[width=0.33\textwidth]{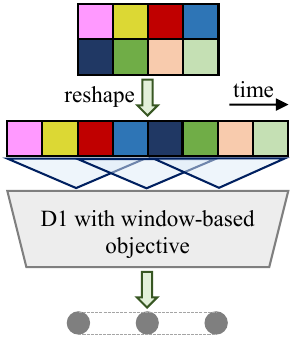}
    \caption{\textbf{Illustration of the reshape operation and the window-based discriminator for our \emph{D2M-GAN}.} 
    }
    \label{fig3:disc}
\end{wrapfigure}
The multi-scale discriminator design has been studied in previous works within the field of audio synthesis and generation~\cite{wang2018high-reso,melgan,hifigan}. 
The discriminator $D = \{D_1, D_2, D_3\}$ in the VQ module of our D2M-GAN is composed of 3 discriminators that operate on the sequence of generated VQ representations and its downsampled features by a factor of 2 and 4, respectively. 
Specifically, different from the multi-scale discriminators proposed in previous works that directly take the raw audio as input, we reshape the VQ representations $f'_{\mathrm{vq}}$ along the temporal dimension before feeding them into the discriminators, which is also important for \emph{D2M-GAN} to reach a stable adversarial training since music is a temporal audio sequence. 
Finally, we use the window-based objectives~\cite{melgan} (Markovian window-based discriminator analog to image patches in~\cite{isola2017image}). Instead of learning to distinguish the distributions between two entire sequences, window-based objective learns to classify between distributions of small chunks of VQ sequences to further enhance the overall coherence as illustrated in Fig.~\ref{fig3:disc}.

\subsection{Lookup and Synthesis}

After generating the VQ representations, we perform a codebook lookup operation similar to other VQ-based generative models~\cite{vqvae,vqvae2,vqgan2021,jukebox} to retrieve the corresponding entries with closest distance. 
Finally, we fine-tune the decoder from the JukeBox~\cite{jukebox} without modifying the codebook entries as the music synthesizer for our learned VQ representations. Specifically, we also adopt the GAN-based technique for fine-tuning the music synthesizer, where the generator is replaced by the decoder of JukeBox and the discriminator follows the similar architecture as described in the previous subsection. 

\subsection{Training Objectives}

\noindent \textbf{GAN Loss.}
We use the hinge loss version of GAN objective~\cite{hinge1,hinge2} adopted for our music generation task to train the proposed \emph{D2M-GAN}.
\begin{equation}
\begin{split}
       L_{adv.}&(D;G)  = \sum_{k} L_{adv.}(D_k;G) \\
        & = \sum_{k} (\mathbb{E}_{\phi(x_a)}[min(0,1 - D_k(\phi(x_a)))] \\
        & + \mathbb{E}_{(x_m,x_v)}[min(0,1+D_k(G(x_m,x_v)))] ),
\end{split}
\end{equation}
\begin{equation}
L_{adv.}(G;D) = \mathbb{E}_{x_m,x_v}[\sum_{k}-D_{k}(G(x_m,x_v))],
\end{equation}
\noindent where $x_a$ is the original music in a waveform, $\phi$ represents the fine-tuned encoder from JukeBox~\cite{jukebox}. $k$ indicates the number of multi-scale discriminators, which is empirically chosen to be 3 in our case.

\noindent \textbf{Feature Matching Loss.} 
To encourage the construction of subtle details in audio signals, we also include a feature matching loss~\cite{feature_matching} in the overall training objective. Similar to the audio generation works~\cite{melgan,hifigan}, the feature matching loss is defined as the $L_1$ distance between the discriminator feature maps of the real and generated VQ features.
\begin{equation}
\centering
\begin{split}
  L_{FM}(G;D) =  \mathbb{E}_{(x_m, x_v)}[\sum_{i=1}^{T}\frac{1}{N_i}\left \| D^{i}(\phi(x_a)) - D^{i}(G(x_m,x_v)) \right \|_1].
\end{split}
\end{equation}

\noindent \textbf{Codebook Commitment Loss.} 
The codebook commitment loss~\cite{vqvae,vqvae2} is defined as the $L_1$ distance between the generated VQ features and the corresponding codebook entries of the ground truth VQ features after the codebook lookup process.
\begin{equation}
\centering
 L_{code}(G) =  \mathbb{E}_{(x_m,x_v)}[\left \| LookUp(\phi(x_a) - G(x_m,x_v)\right \|_1].
\end{equation}

\noindent \textbf{Audio Perceptual Losses.}
To further improve the perceptual auditory quality, we consider the perception losses of the raw audio signals from both time and frequency domains. Specifically, the perceptual losses are calculated as the $L_1$ distance between the original audio and the generated audio samples:
\begin{equation}
\centering
L_{wav}(G) = \mathbb{E}_{(x_m,x_v)}[\left \| x_a - G(x_m, x_v) \right \|_1].
\end{equation}
\begin{equation}
\centering
L_{Mel}(G) = \mathbb{E}_{(x_m,x_v)}[\left \| \theta (x_a) - \theta(G(x_m, x_v)) \right \|_1].
\end{equation}
\noindent where $\theta$ is the function to compute the mel-spectrogram features for the audio signals in waveform.

\noindent \textbf{Final Loss.}
The final training objective for the entire generator module is defined as follows:
\begin{equation}
\centering
\begin{split}
    L_{G} & = L_{adv.}(G;D) + \lambda_{fm}L_{FM}(G;D) + \lambda_{c}L_{code} + \lambda_{w}L_{wav} + \lambda_{m}L_{mel},
\end{split}
\end{equation}
\noindent where the $\lambda_{fm}$, $\lambda_{c}$, $\lambda_{a}$, and $\lambda_{mel}$ are set to be 3, 15, 40 and 15, respectively during our experiments for both levels.

\section{Experiments}

\subsection{Experimental Setup}

\begin{figure}[t]\small
    \centering
    \makebox[0pt][c]{\parbox{1.0\textwidth}{%
    \begin{minipage}[b]{0.49\hsize}\centering
    \includegraphics[width=0.98\textwidth]{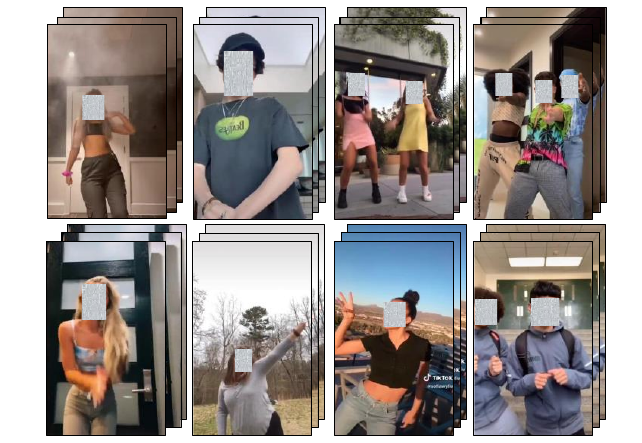}
    \caption{\textbf{Examples of dance videos from our TikTok dance-music dataset.} Different from the AIST dataset~\cite{aist-dance-db} where dancing is performed by professional dancers in a studio environment, our dataset consists of real-world videos collected ``in the wild".}
    \label{fig4:dataset}
    \end{minipage}
    \hfill
    \begin{minipage}[b]{0.49\hsize}\centering
    \includegraphics[width=0.98\textwidth]{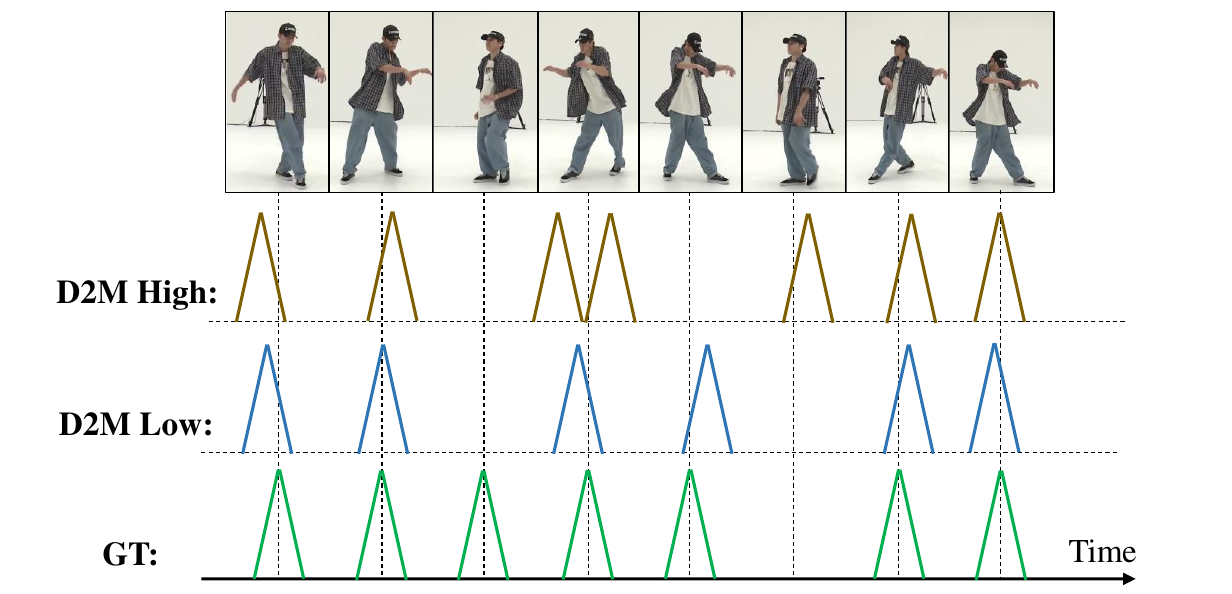}
    \caption{\textbf{Qualitative example of rhythm evaluations and beat correspondence.} The lower-abstraction level model (D2M-Low) appears to align better than its high-counterpart (D2M-High) with the ground-truth (GT), which is consistent with the quantitative scores from the Table~\ref{tab1:results}.}
    \label{fig5:beats}
    \end{minipage}
}}
\end{figure}

\noindent \textbf{Datasets.}
We validate the effectiveness of our method by conducting experiments on two datasets with paired dance video and music: the AIST++~\cite{aistplus} and our proposed TikTok dance-music dataset. 
The AIST++ dataset~\cite{aistplus} is a subset of AIST dataset~\cite{aist-dance-db} with 3D motion annotations. We adopt the official cross-modality data splits for training, validation, and testing, where the videos are divided without overlapping musical pieces between the training and the validation/testing sets. The number of videos in each split is 980, 20, and 20, respectively. The videos from this dataset are filmed in professional studios with clean backgrounds. There are in total 10 different dance genres and corresponding music styles, which include breaking, pop, lock and etc. The number of total songs is 60, with 6 songs for each type of music. We use this dataset for the main experiments and evaluations.
We also collect and annotate a \textbf{TikTok dance-music dataset} which contains 445 dance videos, with an average length of 12.5 seconds. This dataset utilizes 85 different songs, with the majority of videos having a single dance performer, and a maximum of five performers. The training-testing splits contain 392 and 53 videos, respectively, without overlapping songs. Fig.~\ref{fig4:dataset} shows example frames of the dance videos and makes apparent the key differences compared to the professional studio filmed dance video from AIST~\cite{aist-dance-db}. Our videos have wildly different backgrounds, and oftentimes contain incomplete human body skeleton data, which increases significantly the difficulty of the learning problem. For the TikTok music dataset, we use 2D human skeleton data as the underlying motion representation.

\noindent \textbf{Implementation Details.}
For the presented experiments, we adopt a sampling rate of 22.5 kHz for all audio signals. We use the video and audio segments in the length of 2 seconds for training and standard testing in the main experiments. The generation of longer sequences is also investigated in Section~\ref{subsec:ablation}.
The hop lengths for the high and low level are 128 and 32, respectively.
During the GAN training, we adopt the Adam optimizer with a learning rate of 1e-4 with $\beta_1 = 0.5$ and $\beta_2=0.9$ for the generators and discriminators. We define the scaling factor $\sigma = 100$ for the VQ generators. The number of discriminators $k$ is 3 for the multi-scale structure. The batch size is set to be 16 for all experiments.
During the fine-tuning of the JukeBox synthesizer, we use the Adam optimizer with a learning rate of 1e-5 with $\beta_1 = 0.5$ and $\beta_2=0.9$ for the synthesizer and multi-scale discriminators. 
We perform a denoising process~\cite{sainburg2020finding} on the generated raw music data for better audio quality.

\noindent \textbf{Comparisons.} 
We compare our proposed method with several baselines.
\textit{Foley Music~\cite{foley}}: Foley Music model generates MIDI musical representations based on keypoints motion data and then converts the MIDI back to raw waveform using a pre-defined MIDI synthesizer. Specifically, the MIDI audio representation is unique for each musical instrument, and therefore the Foley music model can only generate musical samples with mono-instrumental sound.
\textit{Dance2Music~\cite{aggarwal2021dance2music}}: Similar to~\cite{foley}, the generated music with this method is also monotonic in terms of the musical instrument.
\textit{Controllable Music Transformer (CMT)}~\cite{di2021video}: CMT is a Transformer-based model proposed for video background music generation using MIDI representation. 
In addition to the above cross-modality models that are closely related to our work, we also consider
\textit{Ground Truth:} GT samples are the original music from dance videos.
\textit{JukeBox~\cite{jukebox}}: music samples generated or reconstructed via the JukeBox model. 

\subsection{Music Evaluations}
\label{subsec:evaluation}

We design a comprehensive evaluation protocol that incorporates objective (\textit{i.e.}, metrics that can be automatically calculated) and subjective (\textit{i.e.}, scores given by human testers) metrics to evaluate the generated music from various perspectives. Specifically, the evaluations are divided into two categories: the first category, which is also the focus of our work, measures correlations between the generated music and the input dance videos, for which we compare our proposed model with other cross-modality music generation works~\cite{foley,aggarwal2021dance2music,di2021video} and a random baseline from JukeBox~\cite{jukebox}.
The second category focuses on the quality of the music in general, for which we use the reconstructed samples using JukeBox~\cite{jukebox} given the original audio as input and GT samples for comparisons.

\noindent \textbf{Rhythm.}
Musical rhythm accounts for an important characteristic of the generated music samples, especially given the dance video as input. To evaluate the correspondence between the dance beats and generated musical rhythm, we adopt two objective scores as evaluation metrics, which are the Beats Coverage Scores and the Beats Hit Scores similar to~\cite{lee2019dancing,davis2018visual}. Previous works~\cite{lee2019dancing,davis2018visual} have demonstrated the kinematic dance and musical beats (\textit{i.e.}, rhythm) are generally aligned, we can therefore reasonably evaluate the musical rhythm by comparing the beats from the generated music and those from the GT music samples as shown in Fig.~\ref{fig5:beats}. We detect the musical beats by the second-level onset strength~\cite{ellis2007beat}, which can be considered as the start of an acoustic event. We define the number of detected beats from the generated music samples as $B_g$, the total beats from the original music as $B_t$, and the number of aligned beats from the generative samples as $B_a$. The Beats Coverage Scores $B_g/B_t$ measure the ratio of overall generated beats to the total musical beats. The Beats Hit Scores $B_a/B_t$ measure the ratio of aligned beats to the total musical beats. 
The quantitative results are presented in Table~\ref{tab1:results}. We observe that both levels of our proposed \emph{D2M-GAN} achieve better scores compared to competing methods.

\begin{table*}[t]\small
\centering
\caption{Evaluation protocol and the corresponding results for the experiments on the AIST++ dataset~\cite{aistplus}. \textit{Obj.} stands for \textit{Objective}, which means the scores are automatically calculated. \textit{Subj.} stands for \textit{Subjective}, which means the scores are given by human evaluators.}
\scalebox{0.87}{
\begin{tabular}{cccccc}
\hline
Category   & Features     & Type       & Metric   & Methods    & Scores \\ \hline

 \multirow{5}{*}{Dance-Music} & \multirow{5}{*}{Rhythm} & \multirow{5}{*}{Obj.} & \multirow{5}{*}{\begin{tabular}[c]{@{}c@{}}Beats Coverage\\ \&\\ Beats Hit\end{tabular}}& Dance2Music~\cite{aggarwal2021dance2music}  &  83.5 \&   82.4       \\ \cline{5-5} \cline{6-6}   
 &  &  &   &  Foley Music~\cite{foley}     &  74.1 \& 69.4 \\ \cline{5-5} \cline{6-6} 
  &  &  &   &  CMT~\cite{di2021video}     &  85.5 \& 83.5  \\ \cline{5-5} \cline{6-6}
 &  &  &  & Ours High-level    &   88.2 \& 84.7    \\ \cline{5-5} \cline{6-6} 
   &     &   &  &   Ours Low-level    &   \textbf{92.3 \&  91.7}               \\ \hline \hline
                             
\multirow{5}{*}{Dance-Music} & \multirow{5}{*}{Genre\&Diversity} & \multirow{5}{*}{Obj.} &    \multirow{5}{*}{\begin{tabular}[c]{@{}c@{}}Genre Accuracy \\ (Retrieval-based) \end{tabular}}    & Dance2Music~\cite{aggarwal2021dance2music}  &     7.0     \\ \cline{5-5} \cline{6-6} 
&    &   &  &  Foley Music~\cite{foley}     &  8.1  \\ \cline{5-5} \cline{6-6} 
&  &  &   &  CMT~\cite{di2021video}     &    11.6 \\ \cline{5-5} \cline{6-6}
    &         &          &     &    Ours High-level  &     24.4   \\ \cline{5-5} \cline{6-6} 
      &   &      &  &    Ours Low-level    &  \textbf{26.7}      \\ \hline \hline
                             
\multirow{7}{*}{Dance-Music} & \multirow{7}{*}{Coherence} & \multirow{7}{*}{Subj.} & \multirow{7}{*}{\begin{tabular}[c]{@{}c@{}} Mean Opinion Scores \end{tabular}}& Random JukeBox~\cite{jukebox}  & 2.0 \\ \cline{5-5} \cline{6-6} 
 &    &  &    &     Dance2Music~\cite{aggarwal2021dance2music}     &     2.8    \\ \cline{5-5} \cline{6-6}
         &         &          &    &    Foley Music~\cite{foley}       &       2.8    \\ \cline{5-5} \cline{6-6}
  &  &  &   &  CMT~\cite{di2021video}     &  3.0       \\ \cline{5-5} \cline{6-6}
      &          &        &     & Ours High-level        &        3.5  \\ \cline{5-5} \cline{6-6}
     &        &       &    &    Ours Low-level     &     3.3           \\ \cline{5-5} \cline{6-6}
       &           &      &   &      GT       &            \textbf{4.6}          \\                             
                             \hline \hline
\multirow{4}{*}{Music} & \multirow{4}{*}{Overall quality} & \multirow{4}{*}{Subj.} &  \multirow{4}{*}{\begin{tabular}[c]{@{}c@{}} Mean Opinion Scores \end{tabular}}& JukeBox~\cite{jukebox} & 3.5 \\ \cline{5-5} \cline{6-6} 
   &       &       &     &      Ours High-level     &         3.5              \\ \cline{5-5} \cline{6-6}
  &       &         &    &  Ours Low-level   &           3.7            \\ \cline{5-5} \cline{6-6}
   &         &      &    &  GT   &       \textbf{4.8}          \\                             
                             \hline \hline
\end{tabular}}
\label{tab1:results}
\end{table*}

\noindent \textbf{Genre and Diversity.}
Dance and music are both diverse in terms of genres. The generated music samples are expected to be diverse and harmonious with the given dance style (\textit{e.g.}, breaking dance with strong beats to pair with music in fast rhythm). 
Therefore, we calculate the genre accuracy for evaluating whether the generated music samples have a consistent genre with the dance style. 
The calculation of this objective metric requires the annotations of dance and music genres, we thus use the retrieved musical samples from the AIST++~\cite{aistplus} for this evaluation setting. Specifically, we retrieve the musical samples with the highest similarity scores from the segment-level database formed by original audio samples with the same sequence length. 
The similarities scores are defined as the euclidean distance between the audio features extracted via a VGG-like network~\cite{hershey2017cnn} pre-trained on AudioSet~\cite{gemmeke2017audioset}.
In case that the retrieved musical sample has the same genre as the given dance style, we consider the segment to be genre accurate. The genre accuracy is then calculated by $S_c/S_t$, where $S_c$ counts the number of genre accurate segments and $S_t$ is the total number of segments from the testing split.

We observe in Table~\ref{tab1:results} that the genre accuracy scores of our \emph{D2M-GAN} are considerably higher compared to the competing methods. This is due to the reason that the competing methods rely on MIDI events as audio representations, which require a specific synthesizer for each instrument, and thus can only generate music samples with mono-instrumental sound. In contrast, our generated VQ audio representations can represent complex dance music similar to the input music types, which helps to increase the diversity of the generated music samples. It also makes the generated samples to be more harmonious with the dance videos compared to acoustic instrumental sounds from~\cite{foley,aggarwal2021dance2music}, as shown in the next evaluation protocol for the coherence test.

\noindent \textbf{Coherence.}
Since we generate music samples conditioned on the dance videos, the dance video input and the output are expected to be harmonious and coherent when combined together. Specifically, a given dance sequence could be accompanied by multiple appropriate songs. However, the evaluation of the dance-music coherence is very subjective, therefore we conduct the Mean Opinion Scores (MOS) human test for assessing the coherence feature. During the evaluation process, the human testers are asked to give a score between $1$ and $5$ to evaluate the coherence between the dance moves and the music given a video with audio sounds. The higher scores indicate the fact the tester feels the given dance and music are more coherent.
We prepare the videos with original visual frames and fused generated music samples for testing. In addition to the previously cross-modality generation methods~\cite{aggarwal2021dance2music,foley,di2021video}, we also include the GT samples and the randomly generated music from JukeBox~\cite{jukebox} for comparison.
Our \emph{D2M-GAN} achieves better scores compared to other baselines, which validates the fact that our proposed framework is able to catch the correlations with the given dance video and generates rather complex music that well matches the input.
Details about the human evaluations are included in the supplementary.

\begin{table}[t]\small
\centering
\makebox[0pt][c]{\parbox{1.0\textwidth}{%
\begin{minipage}[b]{0.45\hsize}\centering
\caption{Evaluations for the experiments on the TikTok dataset.}
\label{tab:tiktok}
\scalebox{0.9}{
\begin{tabular}{ccc}
\hline 
Models        & Beats Coverage & Beats Hit  \\ \hline
High w/o M    &    85.5       &    72.4         \\
High w/o V    &     86.3      &     81.7       \\
High (full)   &  \textbf{88.4} & \textbf{82.3}  \\ \hline
Low w/o M &     83.8       &      74.6     \\
Low w/o V &     85.2 &  81.7      \\
Low (full)    & \textbf{87.1}  &  \textbf{83.9}    \\
\hline
\end{tabular}}
\end{minipage}
\hfill
\begin{minipage}[b]{0.52\hsize}\centering
\caption{Results for ablation studies in terms of sequence length.}
\label{tab:ablation_seq}
\scalebox{0.9}{
\begin{tabular}{cccc}
\hline
Length        & Beats Coverage & Beats Hit & Genre Acc. \\ \hline
High - 2s   &    \textbf{88.2}       &   84.7    &   24.4    \\
High - 3s   &   \textbf{88.2}         &   \textbf{85.3}     &    \textbf{25.6}   \\
High - 4s  &  87.1 &  83.0    &  23.3 \\ \hline
Low - 2s &    \textbf{92.3}      &  \textbf{91.7}   &   \textbf{26.7}   \\
Low - 3s &     90.1     &   88.2    &  25.6   \\
Low - 4s& 88.2  & 84.7 &  23.3 \\
\hline
\end{tabular}}
\end{minipage}
}}
\end{table}

\noindent \textbf{Overall Quality.}
Although our main research focus is to learn the dance-music correlations in this work, we also look at the general sound quality of the generated samples. We conduct the subjective MOS tests similar to the coherence evaluation, where the human testers are asked to give a score between $1$ to $5$ for the general quality of the music samples. During this test, only audio signals are played to the testers. The JukeBox samples are obtained by directly feeding the GT samples as input. The MOS tests show that our \emph{D2M-GAN} is able to generate music sample with plausible sound quality comparable to the JukeBox. 
JukeBox has multiple variants with different hop lengths, we compare with samples obtained from the model with same audio hop length for fairness (\textit{i.e.}, the hop lengths for our high and low levels are 128 and 32, respectively). 
It is worth noting that synthesizing high quality audio itself is a vary challenging and computational demanding research topic, for example, it takes \textit{3 hrs} to sample a 20-seconds high-quality music sample with a hop length of 8~\cite{jukebox}.

\noindent \textbf{Results on the TikTok Dataset.}
Compared to the AIST++~\cite{aistplus}, our TikTok dance-music dataset is a more challenging dataset with ``in the wild" video settings that contains various occlusions and noisy backgrounds. 
Table~\ref{tab:tiktok} shows the quantitative evaluation results for the experiments on the TikTok dataset, which demonstrates the overall robustness of the proposed \emph{D2M-GAN}. 

\begin{table}[t] \small
\centering
\makebox[0pt][c]{\parbox{1.0\textwidth}{%
\begin{minipage}[b]{0.48\hsize}\centering
\caption{Results for ablation studies in terms of input modalities on the AIST++ dataset. \textit{M} means the motion data, and \textit{V} means the visual data.}
\label{tab:ablation_modality}
\scalebox{0.78}{
\renewcommand{\arraystretch}{1.21}
\begin{tabular}{cccc}
\hline
Models        & Beats Coverage & Beats Hit & Genre Acc. \\  \hline
High w/o M    & 83.5           & 82.9      & 15.1       \\
High w/o V    & 87.1           & 88.2      & 16.3       \\
High (full)   &  \textbf{88.2} &  \textbf{84.7}  & \textbf{24.4} \\ \hline
Low w/o M & 89.4           & 87.6      & 15.1       \\
Low w/o V & 90.6           & 90.0      & 17.4     \\
Low (full)    &  \textbf{92.3}  &  \textbf{91.7}   & \textbf{26.7} \\
\hline
\end{tabular}}
\end{minipage}
\hfill
\begin{minipage}[b]{0.48\hsize}\centering
\caption{Results for ablation studies in terms of losses on the AIST++ dataset. 
The \textit{mel} loss is especially helpful for beats scores since the beats are characteristic by high frequencies.
}
\label{tab:ablation_loss}
\scalebox{0.75}{
\begin{tabular}{cccc}
\hline
Losses        & Beats Coverage & Beats Hit & Genre Acc. \\ \hline
High w/o $L_{FM}$ & 85.3   &  \textbf{84.7}  &  23.3   \\
High w/o $L_{wav}$ &   85.9       &  \textbf{84.7}  &   23.3     \\
High w/o $L_{mel}$ &    77.6        &   76.5   &    18.6    \\ 
High (full) & \textbf{88.2} &  \textbf{84.7} &  \textbf{24.4}\\
\hline
Low w/o $L_{FM}$ &  91.7  &  90.1  & 24.4    \\
Low w/o $L_{wav}$    &    89.4      &  88.8    &    23.3    \\
Low w/o $L_{mel}$    &   78.8         &     77.1 &    17.4    \\ 
Low (full)      &  \textbf{92.3}   &  \textbf{91.7} & \textbf{26.7} \\
\hline
\end{tabular}}
\end{minipage}}}
\end{table}

\subsection{Ablation Studies}
\label{subsec:ablation}


\noindent \textbf{Sequence Length.}
In the main experiments, we use the 2-second samples for experiments with reference to other similar cross-modality generation tasks~\cite{aistplus}. However, our model can also be effectively trained and tested with a longer sequence length as shown in Table~\ref{tab:ablation_seq} via a larger network with more parameters.

\noindent \textbf{Data Modality.}
We perform ablation studies in terms of the input data modalities, by removing either the dance motion or the visual frame from the input data. Table~\ref{tab:ablation_modality} lists the corresponding experimental results. 
We observe that both motion and visual data contribute to our conditioned music generation task. Specifically, the motion data impose a larger impact on the musical rhythm, which is consistent with our expectations since the musical rhythm is closely correlated with the dance motions.

\noindent \textbf{Loss function.}
We analyze the impact of different losses included in the overall training objective. The results from Table~\ref{tab:ablation_loss} show the contributions of each loss term. Specifically, we observe the audio perceptual loss from the frequency domain $L_{mel}$ helps with the generation of musical rhythm, it is reasonable due to the fact that mel-spectrogram features help to capture the high frequencies from the audio signals, which is closely related to the dance beats. 

\noindent \textbf{Model Architecture.}
We also test various variants of our \emph{D2M-GAN} in terms of the model architecture and proposed model design techniques as in Tabel~\ref{tab:ablation_model}.
The experimental results show that the multi-scale layer for the discriminators, the scaling operation in the generator, as well as the reshape techniques for discriminators are crucial.

\begin{table}[t] \small
\centering
\caption{Results for ablation studies in terms of model architectures on the AIST++ dataset. \textit{D.} means discriminators.
}
\label{tab:ablation_model}
\scalebox{0.8}{
\begin{tabular}{cccc}
\hline
Models        & Beats Coverage & Beats Hit & Genre Acc. \\ \hline
High 1-layer D. &    75.3       &    72.9   &     9.3  \\
High 2-layer D. &       85.3     &    82.9   &   21.0   \\
High w/o scaling  & 72.9 & 71.8 & 14.0 \\
High w/o reshape  & 73.5 & 70.1 & 11.6\\
High w/o fine-tune & 87.0 & \textbf{84.7} & \textbf{24.4}\\ 
High (full) & \textbf{88.2} & \textbf{84.7} &  \textbf{24.4}\\
\hline
Low 1-layer D. &    73.5       &    71.8   &    8.1    \\
Low 2-layer D. &    87.0        &   85.9    &   22.1   \\
Low w/o scaling &  72.4  &  70.1  &  12.8 \\
Low w/o reshape &  73.5  &  71.8 & 12.8\\
Low w/o fine-tune & \textbf{92.3}  &  91.2  & \textbf{26.7}     \\ 
Low (full) &\textbf{92.3} &  \textbf{91.7}&  \textbf{26.7}\\
\hline
\end{tabular}}
\end{table}

\section{Conclusion and Limitations}

To conclude, we propose the \emph{D2M-GAN} framework for complex music generation from dance videos via the VQ audio representations. 
As an early work in the exploitation of VQ based music generation, there are still limitations in the current work from two major aspects: the audio quality and inference speed. As we employ a learning-based encoder-decoder model for raw music (JukeBox~\cite{jukebox}), its performance is the major bottleneck for the quality of our generated music.
Though JukeBox can synthesize relatively high-quality audio signals, there is a tradeoff between computational cost and quality.
Achieving fast inference requires increasing the hop length for the generated waveform, which limits the audio quality and introduces noise.
On the other hand, another direction to balance the above two goals would be investigating a proper approach to automatically compose multiple instruments into a single performance based on video input via MIDI musical representations.


\noindent \textbf{Acknowledgements} This work is partially supported by NSF ECCS-2123521 research grant and Snap unrestricted gift research grant.  This article solely reflects the opinions and conclusions of its authors and not the funding agents.

\clearpage
%
%
\bibliographystyle{splncs04}
\bibliography{eccv2022submissionCR}
\end{document}